\documentclass[accepted]{bmaw2022} 
                                    
\usepackage[american]{babel}
\usepackage{natbib} 
\usepackage{mathtools} 
\usepackage{booktabs} 
\usepackage{tikz} 

\newcommand{\cpr}[2]{\textsf{P}( #1 \, |\, #2 )}
\newcommand{\pr}[1]{\textsf{P}( #1 )}

\title{Redeeming Data Science by Decision Modelling}

\author[1]{\href{mailto:<john-mark.agosta@microsoft.com>?Subject=Your UAI 2022 paper}{John Mark Agosta}{}}
\author[1]{Robert Horton}

\affil[1]{%
    Azure Data\\
    Microsoft\\
    Mountain View CA, USA
}

\begin{document}
\maketitle

\begin{abstract}
  With the explosion of applications of Data Science, the field is has come loose from its foundations. This article argues for a new program of applied research in areas familiar to researchers in Bayesian methods in AI that are needed to ground the practice of Data Science by borrowing from AI techniques for model formulation that we term ``Decision Modelling.'' This article briefly reviews the formulation process as building a causal graphical model, then discusses the process in terms of six principles that comprise \emph{Decision Quality}, a framework from the popular business literature. We claim that any successful applied ML modelling effort must include these six principles.
  
  We explain how Decision Modelling combines a conventional machine learning model with an explicit value model. To give a specific example we show how this is done by integrating a model's ROC curve with a utility model. 
\end{abstract}

\section{ Introduction}

Data Science suffers from its own success, having seen such rapid adoption across so many fields, in so many different ways that it has lost its principled theoretical foundation.  Rational choice as studied in Decision Theory forms a foundation for all analytic fields, which applies no less to Data Science. Many of the ways that Data Science practice falls short should be apparent to anyone versed in this Theory, especially to anyone in our field of Bayesian AI.  The field of Data Science is fluid and evolving rapidly, and defies a concise definition. In contrast mathematical modeling techniques, especially as they have been adopted in Artificial Intelligence, are mature, and can put Data Science on a firm footing. The imperative to better merge Data Science with well understood concepts from Decision Theory should expand the power and scope of its methods.  

At the same time the disruptive ubiquity of software and the scale of data generation in combination with networked hardware platforms---``The Cloud''---creates a new opportunity for AI. 
This fits into a larger social concern such as \emph{ the future of work} (\cite{acemoglu2002technical}) that is drawn into stark relief by the transformation organizations are undergoing due these software innovations are popularly referred to by the term ``Digital Transformation.''

This article argues for a new program of applied research in areas familiar to researchers in Bayesian methods in AI that are needed to ground the practice of Data Science. The article organizes theoretical principles using a framework from the business literature around the list of concepts that comprise \emph{ Decision Quality}. This framework has proved useful in Decision Analysis practice, to connect the practice to the principles that support it.(\cite{Spetzler2016})   We build on these six concepts to clarify the process of building a decision model. 

This article briefly reviews the formulation process as one of building a causal model in Section~2, then discusses the process in terms of Decision Quality, in Section~3, with an example that shows how conventional ROC analysis fits within this framework. For the causal model we use a directed acyclic graph (DAG)---a Bayes network with added decision and value nodes, that goes by various names; ``Influence Diagrams'' (\cite{koller2009}), or ``Decision Graphs'' (\cite{jensen2001}), among others.  We claim that any applied ML modelling effort must include these six principles. In some cases this is obvious, but often it reveals insights into flaws in the model.  Notably those building Data Science models often pay homage to the need to ``understand the business context'' but rarely can explain how to go about it. 
As a specific example of integrating a value model with a predictive model, we show an example of how this can be applied to a predictive model's ROC curve. 

\section{ Formulating a Decision Model }

In simple terms, a machine learning (ML) model predicts an event (Did the customer churn?) or a quantity (What will be product demand next month?) conditioned on a set of observed features. Designate the outcome---the predictive model's target variable $S$, and the vector of features, $\boldsymbol{F}$, the learned model can be written $ \cpr{S}{\boldsymbol{F}}.$ The model provides a distribution over the outcome that ``informs'' a decision made when knowing the features.

The way this is applied is to look up the features $\boldsymbol{f}$ for one case, say the data collected for a customer's purchase history, and compute the probability of outcome $ \cpr{S}{\boldsymbol{f}}$, say their next purchase, as conditional on $\boldsymbol{f}$.  Although referred to as a ``prediction'', an ML model may just as well infer a current unobserved state, such as the root-cause of a failure. In all cases the prediction of the model is an uncertain variable. \emph{One needs to be careful by noting that the model does not predict the \emph{value} of the outcome; value and probability need to be distinguished.}   A value model would express preferences over predicted outcomes and hence it needs to be combined with the prediction to come up with a value. 

\subsection{Decision Modelling}

Once having created a set of alternatives that comprise a decision, 
the combination of a value model with a predictive ML model create the \emph{decision model}. 

A necessary and often overlooked step that precedes the data engineering and model training tasks in Data Science is to properly formulate the model as derived from the decision it is intended to support. We argue for the primacy of starting by identifying the decision in terms of how alternatives interact with values, as opposed to the conventional approach of starting with available data.

A relevant model implies there is an identified action from a set of choices that are predicted to have a desired effect.  If this is not the case then the from an applied point of view, what is the point?  A \emph{decision} refers to making a choice from a set of alternatives, evident as a tangible change at a point in time, in anticipation of the outcomes it precedes. Colloquially one may speak of ``deciding on one's values'', or of thinking of a personal resolution  as a ``decision'' to reform one's behavior.  That's not the sense with which we use the word. However, incidentally,  to resolve one's behavior in such a sense, one may well engage in decision modelling.  We are most  interested when there is uncertainty in the outcomes, that the outcomes of interest by which the best choice will be determined are linked by a chain of cause and effect from the decision to the eventual outcome. 

\subsection{The decision-maker}

Having abstracted the modeling task as one around modeling a decision, there is another abstraction---the question of the \emph{decision-maker}. We apply this term to anchor the model to an individual's choice. ``Individual'' may refer to the person for whom the model is built, or to a class of users, for a decision automated by software, or even for a choice made by an organization.  It is with the \emph{decision-maker} in mind that one
identifies alternatives to be modelled, how the uncertain dynamics play out (the model's predictions), and determines the values of a relevant set of outcomes.  


Once we consider automation, it's no longer a solitary decision, but we are making changes to a decision-making process. The decision could be the response to a recommendation made by the model, as is typical of e-commerce applications. Or it could be a automation of an existing business process. This uncovers a third dimension---of organizational improvements that follow necessarily from the model.

\subsection{Causal models as the canonical formulation tool}

Of course decision variables are not the only ones that make up a model.  The definition of the set of variables that make up the model determine its scope. 
The modeling process begins by setting the scope to ascertain which variables---quantities susceptible to measurement---to include. We partition these into 3 types; \emph{choice} variables that make up a decision, \emph{uncertainties} that describe the world, and \emph{values} that quantify outcomes. There is a first-class distinction between variables that represent
1.) uncertainties as probabilities, 2.) decisions as sets of alternatives, and 3.) outcomes as valued by a utility measure. This partition is both necessary and sufficient to formulate a model.  A glaring lack of most ""plug and play" ML approaches is that they only deal with the probabilistic aspect, and sometimes not even that. 

By formulating influence diagram one creates a structural, causal prior for the model, and defines the inputs and outputs for both the ML and value models.

The decision model can be formulated as an influence diagram from these nodes: 

\begin{itemize}
 
    \item The causal network describing the unobserved state $\boldsymbol{S}.$ These are variables that describe uncertainties relevant to the outcome. 
    \item Variables $\boldsymbol{F}$ that the condition other variables in the model. We partition them into:
    \begin{itemize}
        \item Those that convey information, $\boldsymbol{I}$ meaning they are known when the decision is made, and 
        \item Decision variables $\boldsymbol{d}$; those that are those controlled by the decision-maker. 
        \item Since $\boldsymbol{F} = \boldsymbol{I} \cup \boldsymbol{d}$  both can be inputs to the prediction model. 
    \end{itemize}
    \item The value $v(\boldsymbol{S},\boldsymbol{d})$ is a function of the outcome $\boldsymbol{S}$ and $\boldsymbol{d}$ variables. 
\end{itemize}

\begin{figure}[htp]
    \centering
        \centering
            \includegraphics[width=0.4\textwidth, height=0.24\textwidth]{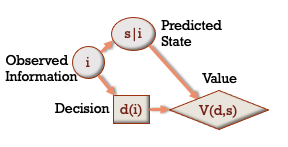}
    \caption{Influence diagram with a predictive model for making an informed decision.}\label{fig:BNwinfo}
\end{figure}

An example of a decision model, Figure~\ref{fig:BNwinfo} shows an influence diagram for the case where the information used to make the prediction is known when the decision is made. A typical example is making a purchase recommendation by knowing a description of the recipient, one makes an informed decision using the prediction $\cpr{S}{i}$. As shown here for a  discriminative ML model the conditioning arc goes from $i$ to $s$. A Bayesian may prefer to learn a generative model with the arc reversed, then apply Bayes rule to solve the diagram.  


In practice the state $\boldsymbol{s}$ may be a network of possibly hundreds of uncertain nodes, connected with a sequence of decision and value nodes, (the decision nodes required to be totally ordered) to form a DAG.  Any influence diagram can be unrolled into a tree, by assuming a total ordering of the DAG, but this soon becomes unwieldy, and the causal claims in the diagram are lost. 

For an application of this influence diagram to an operational example, consider the binary-valued acceptance test of a widget on a production line.\footnote{This example is borrowed from~\cite{Krtuchten2016}.} The decision is to accept widgets inferred to be good and reject those bad. One or more test measurements are made of the widget; this information is used to predict a value by which to classify the state of the widget. The decision is made by minimizing the cost of the relative errors of rejecting a good widget (a false negative) and passing a bad one (a false positive), by thresholding the output of the classifier. The decision rule is simply to reject if the predicted value is less than the threshold and accept otherwise. The rule is determined by ROC (``receiver operating characteristic'') analysis. 

\subsubsection{Causal Models Discovery}

One of the powerful tools in the Bayesian toolbox are network structure learning tools, originating with early work by \cite{heckerman1999} and \cite{spirtes1991}.  Current advances such as "no tears" makes it possible to extend this to continuous variables with non-linear effects (\cite{geffner2022}).
However the placement of the decision and value nodes in the graph are up to the formulation of the decision and not discoverable by learning network structure.

\subsubsection{Optimizing a Decision model}

The influence diagram implies the sequence of computational steps to determine the decision policy $d(\boldsymbol{I})$. 
``Solving'' the decision model of the form shown in Figure~\ref{fig:BNwinfo} reduces to an optimization (here written as maximization),

\begin{equation} V(\boldsymbol{I})^* =  \max_{d(\boldsymbol{I})}E_{\cpr{\boldsymbol{S}}{\boldsymbol{I}}}[V(d(\boldsymbol{I}),\boldsymbol{S})] 
\label{eq:v_subI}
\end{equation}

where each variable may represent a set of nodes in the causal model. Since the decision is made with knowledge of 
$\boldsymbol{I}$ the decision policy becomes $d(\boldsymbol{I})$ and the value function becomes $V(d(\boldsymbol{I}),\boldsymbol{S})$, which becomes a function of the observed features (think a lookup table). Note how machine learning model $\cpr{\boldsymbol{S}}{\boldsymbol{I}}$ is embedded in the decision model.

Written out, the equation says to take expectation over the predictive distribution of the ML model, conditional on the observed features, then for each combination of features make the choice with the highest expected value. 

`Solving" the decision model to determine the best choice means finding the policy $d(\boldsymbol{I})$ that maximizes this expression, given the prediction and value models. This equation applies generally for any data science application.

\section{ Decision Quality Principles }

We discuss each of these principles as they apply to decision models in data science.

\begin{itemize}
        \item\textbf{Create Alternatives} Distinguish decision variables under one's control.
        \item\textbf{Appropriate Frame} Formulate the right problem.
        \item\textbf{Relevant and Reliable Information} Determine the information structure.
        \item\textbf{Clear Values and Tradeoffs} Quantify the utility of outcomes that determine decisions.
        \item\textbf{Sound Reasoning} Apply a valid calculus to solve the model.
        \item\textbf{Commitment to Action} Give ownership to the decision-maker. 
\end{itemize}

In the following sections we define each principle, relate it to the theory that supports it, explain why it is needed, and relate it to an example of the decision to invest in an ML model. 

\subsection{Identify Decisions}
 
In machine learning as in statistics some features are under the decision maker's control -- called \emph {treatments}---others are uncertain characteristics of the environment---sometimes, confusingly called ``controls''. Both make up the features that are inputs to the predictive model.  Confusion of these two kinds of inputs in machine learning models can lead to perverse policies. 

The na\"{i}ve use of predictive models tends to confuse the decision recommendation with the model prediction. An obvious example arises in sales ``propensity scoring'' applications; those that attempt to predict the success of completing a sale based on the product, customer, and economic features. Viewing the prediction as a recommendation confuses the probability generated by the model with the salesperson's decision. As mentioned, in a decision model, the salesperson's decision a variable, in this case an action the salesperson takes to influence the outcome. Consequently in use, salespeople were confused---Does a high ``propensity to close'' mean the sale can be left to its own devices since it's success is inevitable, or does it mean that it needs more to have more effort applied to it? The confusion arises because a propensity model does not include the decision explicitly. 

\begin{figure}[htp]
    \centering
        \centering
            \includegraphics[width=0.4\textwidth, height=0.24\textwidth]{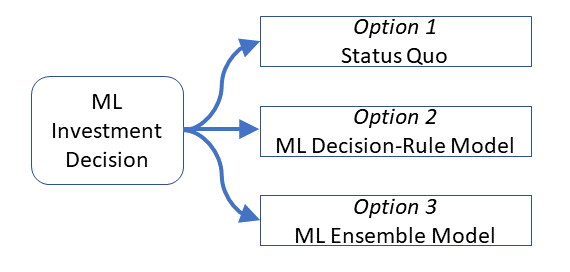}
    \caption{The initial investment node with  a choice of two alternative ML models and the default to do nothing. Option 2 is a simple, low investment choice. Option 3 is more advanced and correspondingly higher investment.}\label{fig:decision_node}
\end{figure}

The example we present demonstrates how the choice of which ML model to apply---one often relegated to irrelevant measures of model performance---can be framed as an investment decision using an influence diagram. We consider the off-line analysis from which two candidate models are built, each described by its ROC curve. These together with the default option, to not use an ML model, make up the three options shown in the summary tree in Figure~\ref{fig:decision_node}.

\subsection{Formulation}

Solving the right problem means not confusing the model and the data with actual phenomenon.  Aside from the obvious question of data quality, by virtue of the causal model one can check if the causal structure, decisions, and value model correspond to the real world. The data often have a physical origin, but the other aspects are derived from subjective factors, often elicited from ``domain experts'' or other problem stakeholders, using techniques borrowed from Human-centered design.  

\begin{figure}[htp]
    \centering
        \centering
            \includegraphics[width=0.5\textwidth, height=0.29\textwidth]{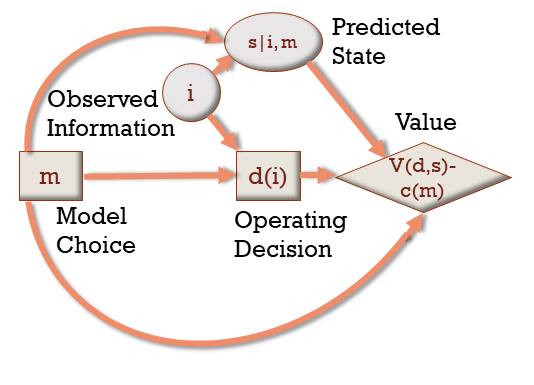}
    \caption{Influence diagram with the added ``offline'' model choice decision.}\label{fig:model_choice}
\end{figure}

In our example, we extend the influence diagram in Figure~\ref{fig:BNwinfo} by pre-pending the model choice investment node $m$ to the diagram shown in Figure~\ref{fig:decision_node}, to create the model investment choice diagram in Figure~\ref{fig:model_choice}. The model choice changes the predictive model $\cpr{\boldsymbol{S}}{\boldsymbol{I}, m}$, by design, includes an investment cost term in the value function. The model choice is an input and hence is known at the time of the operating decision. These three influences are shown by the three arcs that emanate from the investment node. 
Essentially there are three replicates of the previous influence diagram in Figure~\ref{fig:BNwinfo}, each returning a value $V(\boldsymbol{I}, m)^*$. The optimal model choice is simply the maximum over these values:

\begin{equation} V^* = \max_m E_{\pr{\boldsymbol{I}}}[V(\boldsymbol{I}, m)^*]
\label{eq:v_star}
\end{equation}

As a Bayesian representation, the model choice influence diagram requires also that we impute a distribution over the distribution of information to be observed $\pr{\boldsymbol{I}}.$ In common practice this distribution is simply given by the empirical distribution of the model test set, but a complete Bayesian approach allows one to adjust this if its distribution is different in the domain where it is applied. 

\subsection{Relevant Information}

The structure of the decision model can be modified to answer  Value of Information questions. Information value is simply the difference in expected value between the model with and without an arc that conditions a decision variable. If in the ML model the variable's feature importance is negligible, then that conditioning arc will create no information value, however information value brings in an additional consideration: Is the value function sensitive to the state affected by the feature? 

We may consider the model choice variable analogous to a Value of Information choice where the ``information'' is the quality of the ML model employed. Hence we have a method that chooses a model directly on its expected value instead of an indirect measure of accuracy.  We show here how this is done in practice in this example of ROC curve analysis. 

\subsection{Outcome Values} 

There is an inescapable duality between probability and value.  Every predicted outcome has two aspects, a predicted probability of occurrence and its value. A value model maps outcomes into quantifiable values. Utility theory shows us that any consistent set of preferences can be expressed by a utility function. Utility theory, and applications to value modelling as it applies in different domains is a field complementary to probability modelling. 

In the basic case, outcomes can be valued in monetary units to which risk and time preference can be applied. But what if there are multiple outcome variables, each valued differently? For example, I can keep sick patients in the hospital longer to assure their cure, but at the risk of running out of hospital beds if hospital admissions increase.  Perhaps the hardest part of building a value model is the necessity to model trade-offs between competing outcomes, and coming up with a weighting that reduces multiple values to a common scale. The key point is that it is more important to include all factors that determine value, including "intangibles" that require judgment and can not be measured with high accuracy. Better a model that is inclusive instead of a model that avoids important factors presuming they are too hard or subjective to measure. 

Our example illustrates how trade-offs are made in the context of the ``second order'' choice of deciding which predictive model to incorporate in a decision model. The decision maker is the data scientist formulating a decision model; we are applying Decision Quality to the modelling process itself. A flawed practice is to choose just the model with the best test accuracy. This only makes sense in the limit of a perfect model. Otherwise the choice of the model in our example is carried out by ROC analysis. It should depend on the model error rates, which in a binary classfier are two; a \emph{false positive rate} (FPR), and \emph{false negative rate} (FNR).  In addition, the model choice cannot be made without considering the utility tradeoff between the two error rates. This in turn determines a threshold; the operating point with the optimal tradeoff between error rates. Once one estimates the value gained by employing a model (or not), it can be compared with the investment cost for that option, to determine if the option makes sense in total. 

The model choice is made by deriving each model's ROC curves. In a few words, an ROC curve plots the FPR versus FNR for a binary classifier as parameterized by a threshold. The ROC curve is built by running the trained classifier on supervised test data. To determine the optimal operating point one needs a value function expressed as a unit cost for both FPR and FNR errors (and possibly also the costs for correct classification, if not zero). The optimal point---the point with highest utility---occurs where the iso-utility line meets a tangent to the ROC curve. Assuming the ROC curve is convex upward, this point is unique.

\begin{figure}[htp]
    \centering
        \centering
            \includegraphics[width=0.54\textwidth, height=0.34\textwidth]{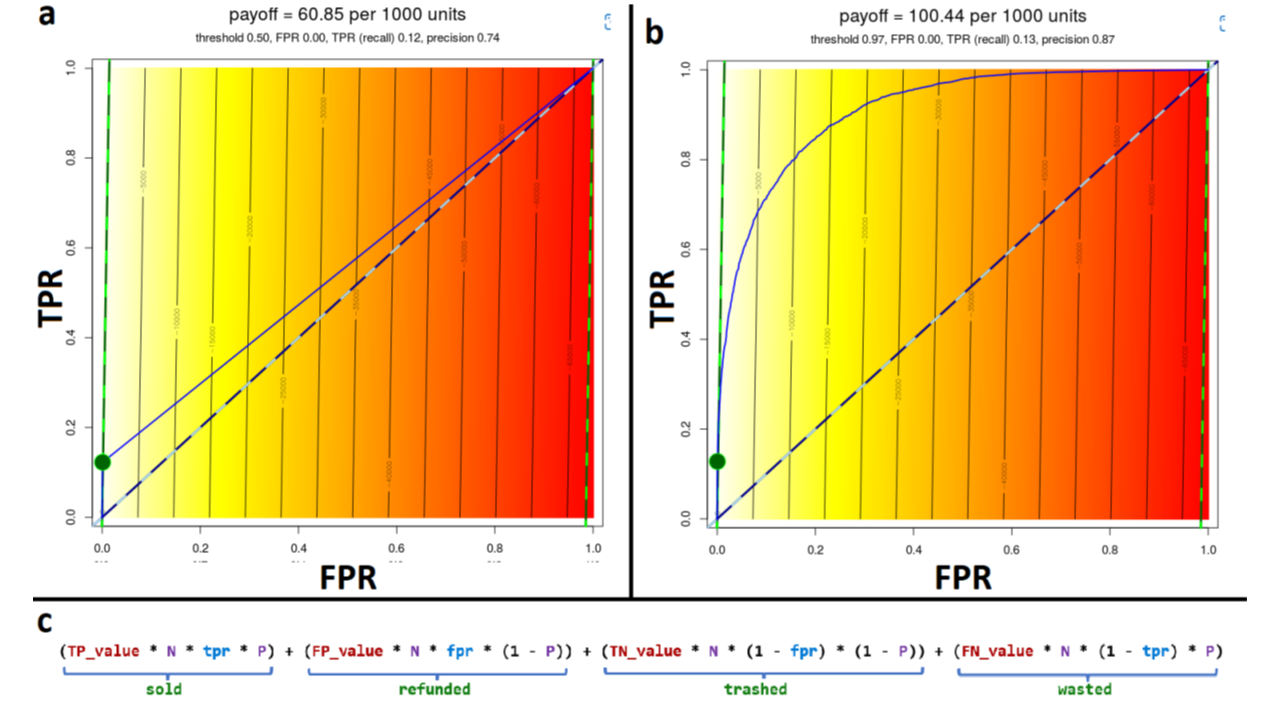}
    \caption{ROC curves that express the value of using each of the two models.}\label{fig:2roc}
\end{figure}

The two panels in Figure~\ref{fig:2roc} show the same ROC curve for the two different predictive models. The colored background shows the utility for each point in the background. The point on the ROC curve where the utility is highest is indicated by a green spot, and the payoff at that point is shown above the plot, as are the model score threshold, the true positive rate (TPR) and false positive rate (FPR). 

The image in panels (a) and (b) are screenshots from our R Shiny app\footnote{https://ml4managers.shinyapps.io/ML\_utility/}, made using the following inputs:

\begin{tabular}{|c|c|c|}\hline
    \textbf{Variable}& \textbf{Description}& \textbf{Value}\\\hline
	 P& proportion positive& 0.02\\
	 TP\_value& value of true positive& 50\\
	 FP\_value& value of false positive& -70\\
	 TN\_value& value of true negative& 0\\
	 FN\_value& value of false negative& 0\\\hline
\end{tabular}

Under these conditions, the highest value on the diagonal is zero, at the origin. This means that without a way to select widgets with above average probability of being good, the expected value of selling widgets is negative, and your best bet is to not sell any (giving an expected value of zero for the ‘status quo’, Option 1). The line of indifference for the value at the origin is indicated by the dashed green and black line on the left edge of the figure. Only ML models whose ROC curves cross this line have value greater than not using an ML model at all; note that in these example our ROC curves do cross the line, but just barely.

Panel (a) shows the ROC curve for a simple binary rule based on a single data feature (Option 2); cases meeting this rule have a high proportion of positives, but only a small fraction of the good widgets are detected this way. Such rules may have low implementation costs, however.

Panel (b) shows the ROC curve for a more sophisticated ML model (Option 3, in this case a Random Forest model). The value is slightly higher than the simple rule of option 2, but be aware that because this model depends on multiple data features, its cost of implementation and operation are expected to be higher as well.

The choice of model is simply the model---possibly among several--- with the highest utility at it's optimal threshold as determined by the ROC analysis. This is equivalent to computing the expected value of a model by Equation~(\ref{eq:v_subI}). Then the choice of the model follows by Equation~(\ref{eq:v_star}). This is the expected utility of a decision using that model in the investment decision model. 

{\bf The important thing to note is that the evaluation and thus choice of the predictive model depends strongly on the utility function that applies to its errors as well as on any intrinsic property of the model.}  For anything short of a perfect model there is no one best model; one model may be better when its FPR is less costly and vice versa. Furthermore, even a model with a higher expected value must be compared based on it's development and operational costs.

ROC model analysis is part of the off-line predictive model development task, preceding the implementation of the operational decision model, however the formulation of the operational model, specifically its value function must be known for the off-line analysis.  Since the best model choice depends on factors that are not intrinsic to the model, i.e. the utilities, and also the  base rate ``prevalence'' of the condition to be classified, one could imagine automating the predictive model selection as conditions change in the primary model. 

This model selection framework can be extended to classification and regression problems in general. An ROC value model has a natural generalization to multi-valued outcomes as presented in \cite{landgrebe2008}. The optimization and inference steps will change as the problems change, but their combination as shown by the influence diagram~\ref{fig:model_choice} applies uniformly---the influence diagram that expresses the model investment decision problem does not change.

\subsection{Sound Reasoning} 

As Bayesians we tend to have a good handle on sound solution methods. Much of the Bayesian literature among statisticians argues for coherent probabilistic reasoning. By the nature of the Bayesian program we are well equipped to assure that a model's claims are valid.   One point of contention may be the idea that ``truth'' in conventional ML practice derives from the testing on a holdout set of data from the set trained on---implicitly this assumes that the model prior distributions are given by the test set empirical distribution.  The Bayesian approach offers a way to adjust priors should these domain distributions shift over time. 

\subsection{Commitment}

Behavioral psychology (\cite{kahneman2011}) explains why people make irrational choices when outcomes are uncertain and far off. The harder question is often how to create commitment despite people's natural tendencies. 

Behavioral aspects that determine whether a model is put into practice and its recommendations accepted are of course necessary for its value to be realized.  How such commitment is assured, or equivalently what does it take for a model to be accepted brings in a host of concerns outside the Bayesian program. Making a decision has a human, emotional side. Often what is lacking is the decision-maker's understanding of the model; its interpretability. Having a causal basis, as expressed by the influence diagram structure of the model not only makes the model more interpretable, but extends the interpretation to explanations of what in the real world is modelled, not just an interpretation of how the model functions. 


\section{Conclusion} 

The field of decision modelling as an outgrowth of data science and decision theory suggests a program of research that is in its early days.  We have proposed an approach that extends conventional ML practice by using influence diagrams to create integrated predictive and value models. We gave an example of making a model selection choice for a binary classifier with a linear utility function. Future work will extend this to other predictive models, and their integration with more general utility models. 

Data science is now beset by a host of thorny ethical questions about ``responsible AI''. Perhaps the field would be advanced by a  ``reverse'' modeling approach, with models that recover the decision-maker's true preferences, as proposed in Stuart Russell's book \emph{Human Compatible}.\cite{Russell2019}



\bibliography{uai2022-ds}

\appendix

\end{document}